# Using virtual human for an interactive customer-oriented constrained environment design

Liang MA , Ruina MA, Damien CHABLAT , Fouad BENNIS

IRCCyN, UMR CNRS 6597
Ecole Centrale de Nantes,
1 Rue de la Noë,
44321 Nantes Cedex 3, France
*Phone: (+33)(0)240376958*
*E-mail* : {liang,ma, ruina.ma, damien.chablat, fouad.bennis}@irccyn.ec-nantes.fr

**Abstract:** For industrial product design, it is very important to take into account assembly/disassembly and maintenance operations during the conceptual and prototype design stage. For these operations or other similar operations in a constrained environment, trajectory planning is always a critical and difficult issue for evaluating the design or for the users' convenience. In this paper, a customer-oriented approach is proposed to partially solve ergonomic issues encountered during the design stage of a constrained environment. A single objective optimization based method is taken from the literature to generate the trajectory in a constrained environment automatically. A motion capture based method assists to guide the trajectory planning interactively if a local minimum is encountered within the single objective optimization. At last, a multi-objective evaluation method is proposed to evaluate the operations generated by the algorithm.

**Key words**: virtual human, constrained environment design, optimization, motion capture, trajectory planning.

## 1- Introduction

For industrial products, a compact design decreases the required massive space and enhances the appearance of the product. From another aspect, the designers have to consider constrained environments resulting from the compact design. Under constrained situations, assembly/disassembly oriented design has to be taken into consideration, since there are several ergonomic issues for the end user of the product or for the maintenance process. For example, the visibility [SM] and accessibility [LD, RS] of a component during an assembly operation; awkward posture caused by the product layout; physical or mental fatigue from the operations, etc.

For these reasons, virtual human simulations are often engaged during the conceptual design stage to evaluate the accessibility of the virtual environment and other ergonomic aspects [MC1, RM]. Thanks to the interaction between the virtual human and the constrained environment, the designers are able to evaluate the manual handling operations, plan the possible trajectories, and further improve the design.

Trajectory planning is one of the most important problems for the use of virtual human in product design. In general, three approaches have been used frequently in the literature to generate the trajectory: inverse kinematics, optimization-based method, and motion capture method. Inverse kinematics can generate a trajectory automatically and rapidly; however, this method could not generate a collision-free path easily. In order to overcome this inconvenience, an optimization-based approach has been proposed in [RM] to find a collision-free path iteratively. In comparison to inverse kinematics, direct kinematics has been used in the optimization based approach and it enhances the computation efficiency. However, sometimes, the path can be trapped in a local minimum and it cannot get out from it without external intervention. Using motion capture method, it is convenient to achieve natural movement in a virtual environment. However, the motion data obtained from the motion capture has to be processed using motion retargeting method to adapt it to the overall population.

In our research, we are aiming at creating a trajectory planning and evaluation method to improve the product design during the conceptual design stage. Virtual human modelling is taken to represent the overall population with different anthropometrical data. A single objective optimization based method is used to generate the trajectory at first. Then, motion capture methods or other intervention methods are used to help the algorithm to move out from the local minimum. At last, multi-objective evaluation methods are going to be used to evaluate the generated trajectory.

## 2- Trajectory planning algorithm

A virtual human is modelled using the modified Devanit-Hartenberg method [KD] with 28 degrees of freedom (DOF) to describe the mobility of all the key joints around human body. The kinematic information could be described by a set of general coordinates $[\mathbf{q}, \dot{\mathbf{q}}, \ddot{\mathbf{q}}]$, where **q** is the set of the rotational angles representing the positions of each joint [MZ1].

### 2.1 – Single objective optimization

#### 2.1.1 – Trajectory planning algorithm

```
SOO based algorithm 1
Get current posture q
Initialize the trajectory Traj by adding q to tail
Current position of the end effector P_end=direct_kinematic(q)
Distance D=Dist(P_end, P_des)
while D>ε
    1. Generate a set of new postures in
    the neighbourhood (q±dq)
    2. Select the collision-free postures
       Calculate the distances Dist_set
    3. Find out the nearest collision-free posture
       q=min(Dist_set)
    4. Update current posture q
    5. Update the distance D
    6. Add the current new posture q to trajectory Traj
end while
```

**Figure 1: Single objective optimization based trajectory planning method**

This single objective optimization (SOO) based method was proposed by [RM] in order to generate the trajectory automatically, and its principle is shown in Fig. 1. In this algorithm, the distance between the end effector and the destination is chosen as an objective function. For a virtual human, the position of the destination is known and the current posture **q** is also known. A change (**±dq**) to the current posture configuration is added to obtain several posture candidates for the next movement. Candidates without collision to the virtual environment are selected out via collision test. The one among the rest candidates with the smallest distance is selected to update the current posture.

#### 2.1.2 – Technical problem

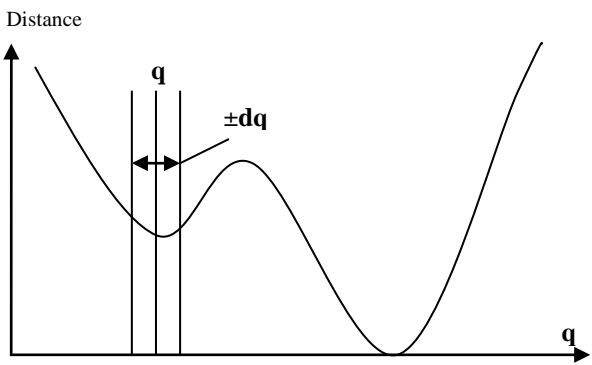

**Figure 2: Local minimum encountered in single objective optimization based method**

One of the greatest technical problems in this method is the local minimum encountered while searching the direction to the destination. This problem is illustrated by a simple example in Fig. 2. During the trajectory planning, it is very possible that the optimization process will encounter the local minimum. In this case, this algorithm will be trapped and cannot advance anymore to its global minimum. In this case, the step length **dq** can be modified to skip the local minimum, or the configuration **q** can be changed by another posture configuration. These modifications need to be done using external intervention.

### 2.2 – Trajectory planning via external intervention

#### 2.2.1. – Modified algorithm

As what has been discussed in the section 2.1, the single objective optimization method cannot avoid local minimum and that results in no evolution for finding a trajectory to the destination. Therefore, a modified algorithm is proposed in this section using external intervention to overcome this difficulty. The algorithm is presented in Fig.3. Since the step length is constant without intervention, if a posture **q** has appeared again in the trajectory, there comes local minimum in the trajectory.

```
Modified SOO based Algorithm
Get current posture q
Initialize the trajectory Traj by adding q to tail
Current position of the end effector P_end=direct_kinematic(q)
Distance D=Dist(P_end, P_des)
while D>ε
    1. Generate a set of new postures in
    the neighbourhood (q±dq)
    2. Select the collision-free postures
       Calculate the distances Dist_set
    3. Find out the nearest collision-free posture
       q=min(Dist_set)
    if q exists in the trajectory (local minimum) and D>ε
        Get a new q from external intervention
    end if
    4. Update current posture q
    5. Update the distance D
    6. Add the current new posture q to the trajectory Traj
end while
```

**Figure 3: Modified SOO based method**

#### 2.2.2. – Technical problems

An external intervention is implemented via different methods. In this part, a multi-agent thought is introduced into our system. The thought is explained by Fig. 4.

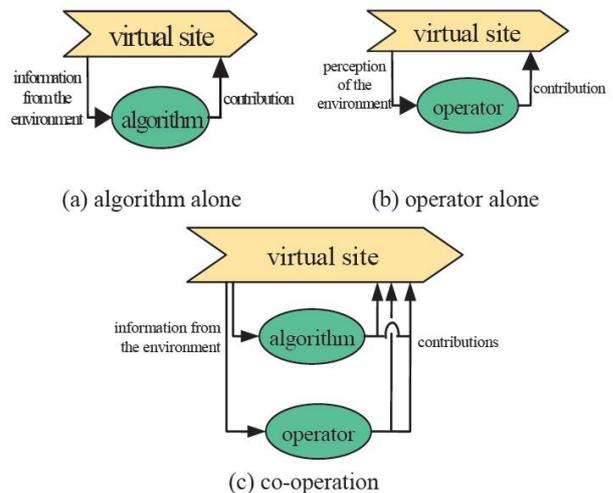

**Figure 4: Co-operation principles**

In a virtual space, just using algorithm along could plan a path for the virtual human (Fig. 4(a)). At the same time, operating the virtual human directly by changing its rotational configuration manually could also avoid the obstacles to get to the target position (Fig. 4(b)). Each way has its disadvantages: using Fig. 4(a) will encounter the local minimal obviously; using Fig. 4(b) is difficult to generate human-like action and much more time consuming. For these reasons, the co-operator principle (Fig. 4(c)) is implemented by us to combine the advantages of each way to get a better result [CC].

In order to solve these problems, the direct method is to provide a graphical interface to change the configuration: the posture **q** or the step length **dq**. Through changing these parameters the mannequin can be lead to go out of the local minimum situation. Another solution is using motion capture method to skip the local minimum.

In the first solution, changing the posture **q** can guide the mannequin out of the local minimum situation, but it should be known that it is difficult to have an intuitive manipulation to operate the virtual human by changing the rotational configuration. It requires much more time to manipulate the angular variables directly. Enlarging the step length **dq** can also guide the mannequin to move out the local minimum situation, but the path will not be so smooth.

In the second solution from a motion tracking system, different motion data of a human body can be obtained. The virtual human could be operated much more naturally to skip the local minimum. In the second method, a motion tracking system is required. Accompanying with this method, inverse kinematics or motion retargeting techniques have to be developed to map the motion data to the simulated trajectory.

The main problem is now to define the algorithm which is able to use the principle of the multi-agent system [CC] and to add the information of the motion capture system.

### 2.3 – Multi-objective evaluation

#### 2.3.1. –Ergonomic objectives for evaluating the constrained environment

As we have mentioned in Introduction, there are different aspects that the designers have to respect. To produce a well designed constrained environment, visibility and accessibility are both important factors. Besides them, the physical influence from the environment should also be assessed in some cases. Therefore, a multi-criteria evaluation system is proposed in this section to evaluate the constrained environment.

**Accessibility**: this term describes whether the user could obtain an access to a certain component in the environment. It could be evaluated by the number of possible trajectory solutions (*N*). The larger the number of solutions, the easier the component can be accessed. If there is no solution for the trajectory, the component is not accessible by a human being.

**Visibility**: this term describes the visual accessibility of a component. This term has been modelled or used to analyze workspace in the literature [CC, MJ, SM]. In our research, the visibility is going to be integrated into trajectory planning by treating it as one of the end effectors, since the visible region is also an important factor determining the feasibility of the operation.

**Posture effect**: this term describes the effect resulting from posture. This is a traditional subject in ergonomic analysis, and there are several conventional methods for evaluating the posture [MA, ME]. During the ergonomic application, duration of the task, posture engaged in the task, and its physical exposures are taken to evaluate the potential risks of the posture.

**Fatigue**: this term is used to describe the effect of physical load on the human body. It has been modelled in [MC2] according to physical exposures related to the manual handling operations. This term is used in our research to evaluate the physical effects of the task realized in the constrained environment.

#### 2.3.2. –Technical approach

As discussed before, multi-criteria evaluation system is going to be established to assess different ergonomic aspects of a constrained environment. In this approach, different aspects are mathematically modelled to create objective evaluation. Those results could be useful to improve the design of constrained environment.

Meanwhile, a multi-objective optimization procedure could also be interesting to determine design parameters of a constrained environment. This algorithm is presented in Fig. 5.

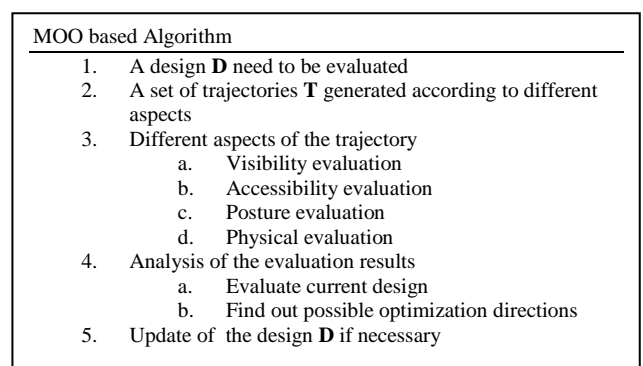

MOO based Algorithm
1. A design **D** need to be evaluated
2. A set of trajectories **T** generated according to different aspects
3. Different aspects of the trajectory
   a. Visibility evaluation
   b. Accessibility evaluation
   c. Posture evaluation
   d. Physical evaluation
4. Analysis of the evaluation results
   a. Evaluate current design
   b. Find out possible optimization directions
5. Update of the design **D** if necessary

**Figure 5: Multi-objective optimization based evaluation and trajectory planning algorithm (MOO)**

### 3- Application case

#### 3.1 – Systems

In order to realize our algorithm, a virtual reality platform is constructed. This platform includes two parts: simulation system and motion capture system. The simulation system is mainly responsible for the generation of virtual environment, the collision computation, and the automatic trajectory planning. An optical motion tracking system is in charge of capturing motion data and communicating with the simulation system.

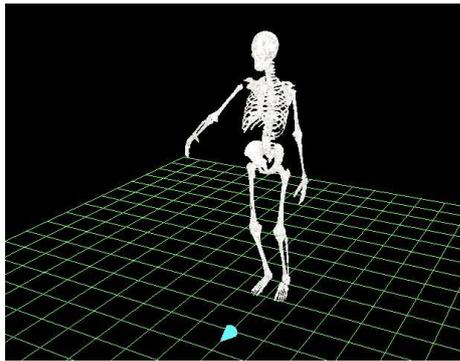

**Figure 6: The virtual skeleton in the simulation system**

The simulation system is developed using OpenGL and C++. The virtual skeleton is shown graphically in Fig. 6. The virtual human is combined by ten body parts: head, torso, thighs, shanks, upper arms, and lower arms. Each body part is modelled as a 3DS model file which is composed of hundreds of triangle facts. The virtual skeleton is driven using direct kinematic method by changing angular values of each key joints. Meanwhile, virtual environments could also be loaded from 3DS files which are converted directly from CAD models.

In the motion capture system (Fig 7), there are totally eight CCD cameras to capture the motion in a range of 2mx2mx2m. Nonlinear direct transformation method is used to calibrate all the cameras. After calibration, the system could capture maximum 13 markers at 25Hz [MZ2]. Although this tracking speed is not enough for capturing accurate motion, it still provides an acceptable speed to adjust the posture.

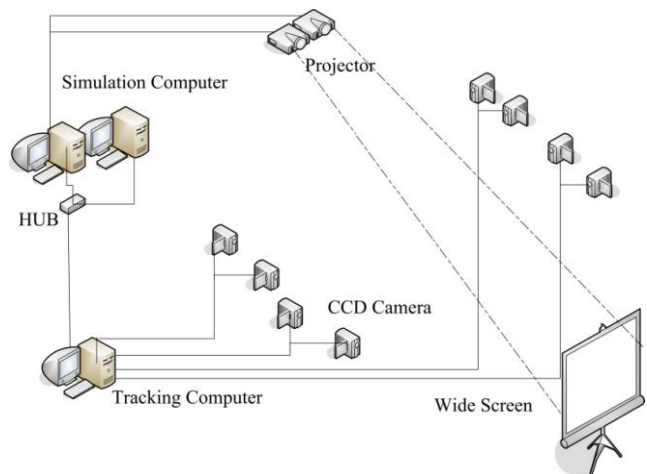

**Figure 7: Motion Capture System**

### 3.2 – Robot trajectory planning

Since there are too many DOFs in a virtual human, at the very beginning of our research, a trial demonstration of the algorithm has been realized by using a RRR robot and several virtual objects in a virtual environment (see Fig. 8-10). The robot is composed of three rectangles, and the end effector is the right end. The blue round point denotes the destination of the end effector. There are three obstacles (two triangles and one block) in the environment.

Different trajectories will be generated by using our algorithm according to the different length of each revolute joint in the RRR robot. Fig 8 shows that trajectory of RRR robot with the link length parameters (20, 10, 20). Fig 9 shows that the trajectory of RRR robot with the link length parameters (20, 20, 25) and Fig 10 shows that the trajectory of RRR robot with the link length parameters (20, 20, 40). From these figures, we can see that in the same environment, different size of the RRR robot can come across various situations. This is necessary and also important, because of a product is not just for a fixed user. Various situation or parameters of a subject should be taken into consideration. The lengths of the links are changed demonstrate different effects of dimensional information.

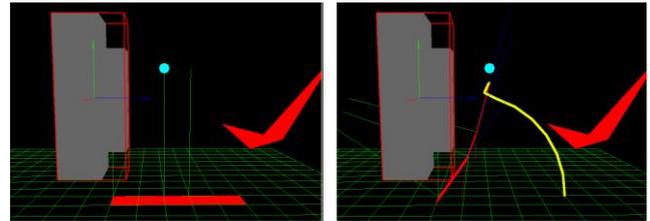

**Figure 8: Trajectory planning test using a RRR robot (with link length equals 20, 10, 20)**

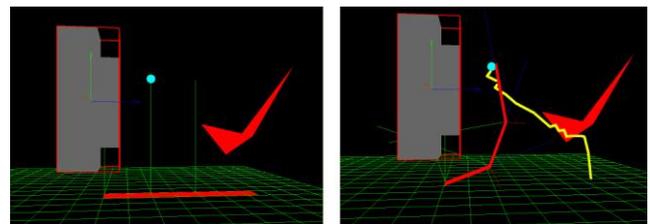

**Figure 9: Trajectory planning test using a RRR robot (with link length equals 20, 20, 25)**

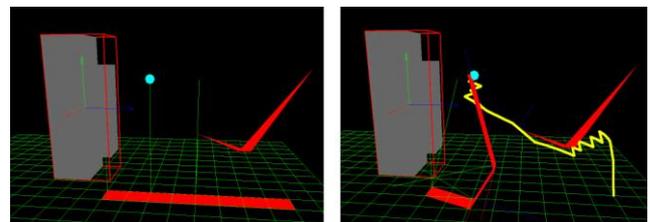

**Figure 10: Trajectory planning test using a RRR robot (with link length equals 20, 20, 40)**

The trajectory in yellow is generated by the algorithm presented in section 2.1.1. It is observable that the trajectory in yellow could avoid the collision while approaching to the destination. Fig 8, 9 and 10 show that with the same obstacle different geometrical configurations (different arm lengths) can generate different paths to avoid the obstacle from the same start to the same destination.

In the obstacle avoiding process, there is always possibility of local minimum. In Fig. 11, a demonstration of local minimum is shown.

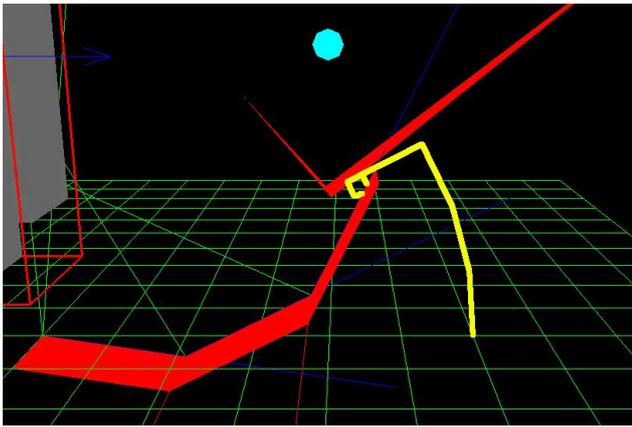

**Figure 11: Trajectory planning test using a RRR robot (with local minimum, step=0.08)**

Since the obstacles locate between the destination and the end effector and the descending direction is also restricted by the obstacles, the algorithm could not skip the local minimum with a step length 0.08.

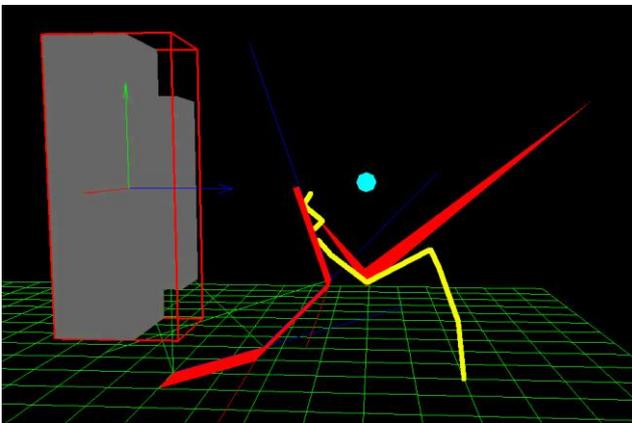

**Figure 12: Trajectory planning test using a RRR robot (External intervention interface, step=0.1)**

In Fig.12, for the same arrangement of obstacles in Fig 10, the step length has been adjusted to 0.1. As a result, the first obstacle could be passed over without problem.

### 3.3 – Virtual human trajectory planning

Virtual human trajectory planning using the proposed algorithm is still under construction. There are still several steps to complete the demonstration: motion retargeting, modelling of different aspects, and the complete installation of the motion capture system.

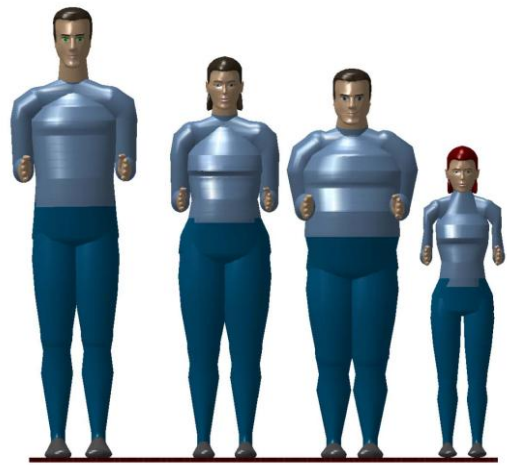

**Figure 13: Set of mannequins to be used to test new products**

The definition of the trajectories should be not just used for one person, because of a product is designed for a given population (Fig. 13). An automatic path planner can calculate the path from a start point to a destination. But imaging that in a complex environment and for many users, sometimes the algorithm might be failed. In this situation, the user interaction has to be limited to minimize design effort.

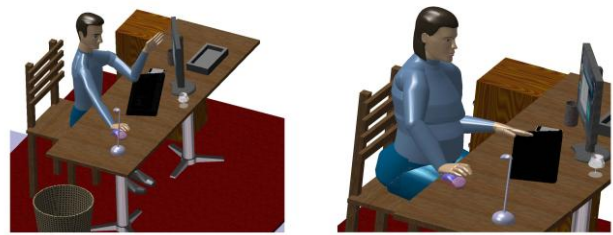

**Figure 14: Example of office design for two different women**

The ergonomic study of the product can be tested thanks to the definition of a set of tasks: taking a mouse, touching the screen (Fig. 14). Different tasks have different aspects to be respected, and different aspects will result different trajectories.

Different sizes, different weights, different tasks, and different design aspects: all these factors lead trajectory planning for a virtual human to a quite difficult problem. The multi-objective evaluation and optimization approach for virtual human trajectory planning has to be developed with caution.

## 4 Conclusions and perspectives

In this paper, an approach of using virtual human in constrained environment design was presented, and a new algorithm was proposed to solve trajectory planning. Our preliminary application of the algorithm demonstrates that this interactive method had a potential to help us solve trajectory planning problem in constrained environment. In our future research, a multi-agent system and multi-objective optimization method will be implemented to facilitate the design process of constrained environments.

## 5- Acknowledgements

This research is supported in the context of collaboration between the Ecole Centrale de Nantes (Nantes, France) and Tsinghua University (Beijing, P.R.China). The lead author would also like to thank Ecole Centrale de Nantes for the financial support of the post-doctorate studies. The authors would like to address thanks to the financial support of Ecole Centrale de Nantes, Région des Pays de la Loire in the project Virtual Reality for design (VR4D) and China Scholarship Council (CSC).

## 6- References

**[CC]** Chedmail, P.; Chablat, D. & Roy, C. L. A distributed approach for access and visibility task with a manikin and a robot in a virtual reality environment IEEE Transactions on Industrial Electronics, 50, 692-698, 2003

**[LD]** Li, K.; Duffy, V. & Zheng, L.Universal accessibility assessments through virtual interactive design International Journal of Human Factors Modelling and Simulation, Inderscience, 1, 52-68, 2006

**[MA]** MA, L.PhD Thesis : Contributions pour l'analyse ergonomique de mannequins virtuels. Ecole Centrale de Nantes, 2009.

**[MC1]** B. Maille, P. Chedmail and E. Ramstein, Emergence of a virtual human behaviour in a virtual environment, 4ème Conférence Internationale sur la Conception et la fabrication Intégrées en Mécanique, IDMME, Clermont-Ferrand, France,Mai,2002.

**[MC2]** Ma L., Chablat D., Bennis F. and Zhang W., "A New Simple Dynamic Muscle Fatigue Model and its Validation", International Journal of Industrial Ergonomics, Vol. 39(1), January 2009, pp. 211-220.

**[ME]** McAtamney, L. & Corlett, E. N.RULA: A survey method for the investigation of work-related upper limb disorders Applied Ergonomics, 24, 91-99, 1993

**[MJ]** Masih-Tehrani, B. & Janabi-Sharifi, F.Kinematic modeling and analysis of the human workspace for visual perceptibility International Journal of Industrial Ergonomics, Elsevier, 2008(38), 73-89

**[MZ1]** Ma, L.; Zhang, W.; Chablat, D.; Bennis, F. & Guillaume, F. Multi-objective optimisation method for posture prediction and analysis with consideration of fatigue effect and its application case. Computers & Industrial Engineering, Elsevier, 2009(57) 1235-1246.

**[MZ2]** Ma, L.; Zhang, W.; Fu, H.; Guo, Y.; Chablat, D. & Bennis, F. A Framework for Interactive Work Design based on Motion Tracking, Simulation, and Analysis. Human Factors and Ergonomics in Manufacturing & Service Industries, 2010

**[SM]** Smith, B.; Marler, T. & Abdel-Malek, K.Studying visibility as a constraint and as an objective for posture prediction, SAE International Journal of Passenger Cars - Mechanical Systems, SAE International, 2009(1), 1118-1124

**[KD]** Khalil, W. & Dombre, E.Modelling, identification and control of robots Hermes Science Publications, 2002

**[RM]** A. Rennuit, A. Micaelli, X. Merlhiot, C. Andriot, F. Guillaume, N. Chevassus, D. Chablat and P. Chedmail, Passive Control Architecture for Virtual Humans, International Conference on Intelligent Robots and Systems,Edmontgon, Canada,2005.

**[RS]** Rajan, V.; Sivasubramanian, K. & Fernandez, J. Accessibility and ergonomic analysis of assembly product and jig designs. International Journal of Industrial Ergonomics, 1999(23), 473-487